\documentclass{article}

\usepackage{PRIMEarxiv}
\usepackage[utf8]{inputenc} 
\usepackage[T1]{fontenc}    
\usepackage{hyperref}       
\usepackage{url}            
\usepackage{booktabs}       
\usepackage{amsfonts}       
\usepackage{nicefrac}       
\usepackage{microtype}      
\usepackage{lipsum}
\usepackage{fancyhdr}       
\usepackage{graphicx}       
\graphicspath{{media/}}     
\usepackage{amsmath}
\usepackage{todonotes}
\usepackage{array}          
\usepackage{adjustbox}      
\usepackage{makecell}
\usepackage{pgfplots}
\usepackage{pgfplotstable}
\pgfplotsset{compat=1.18}
\usepackage{longtable}
\usepackage{geometry}
\usepackage{blindtext}
\usepackage[numbers,square,comma]{natbib}

\usepackage{listings}
\title{Are Frontier Large Language Models Suitable for Q\&A in Science Centres?}

\author{
Jacob Watson, Fabrício Góes, Marco Volpe, Talles Medeiros \\
Computing and Mathematical Sciences Department \\
University of Leicester \\
Leicester, UK\\
\texttt{\{jw974,fabricio.goes,mv163,thm14\}@leicester.ac.uk
}}

\begin{document} 
\maketitle

\begin{abstract}
This paper investigates the suitability of frontier Large Language Models (LLMs) for Q\&A interactions in science centres, with the aim of boosting visitor engagement while maintaining factual accuracy. Using a dataset of questions collected from the National Space Centre in Leicester (UK), we evaluated responses generated by three leading models: OpenAI's GPT-4, Claude 3.5 Sonnet, and Google Gemini 1.5. Each model was prompted for both standard and creative responses tailored to an 8-year-old audience, and these responses were assessed by space science experts based on accuracy, engagement, clarity, novelty, and deviation from expected answers. The results revealed a trade-off between creativity and accuracy, with Claude outperforming GPT and Gemini in both maintaining clarity and engaging young audiences, even when asked to generate more creative responses. Nonetheless, experts observed that higher novelty was generally associated with reduced factual reliability across all models. This study highlights the potential of LLMs in educational settings, emphasizing the need for careful prompt engineering to balance engagement with scientific rigor.
\end{abstract}

\section{Introduction}\label{introduction}

Large Language Models (LLMs) such as GPT (Generative Pre-Trained Transformer), Claude and Gemini have advanced considerably in recent years, now being capable of maintaining conversations and providing complex and creative answers to prompts. These frontier models provide opportunities to science centres and museums to capture audiences and enhance the experience of visitors through emergent and procedural experiences, educating them about the institution's chosen thematic field while ensuring the delivery that is entertaining, engaging and creative \cite{Gemeinhardt2024,vasic2024,Hettmann2023}. However, LLMs must be used with caution to mitigate well-known risks such as hallucinations, bias and misinformation \cite{yan2023, georgios2023}. 

In educational contexts like science centres, LLMs have shown promise in both enhancing accuracy and public engagement through interactive experiences. For instance, approaches such as TUTOREVAL and TUTORCHAT have bolstered models' accuracy in science tutoring \cite{vasic2024}, while prompt engineering alone has yielded high performance on scientific benchmarks like SciQA \cite{yan2023}. Other applications include personalized tours in virtual environments \cite{georgios2023} and interactive Q\&A systems providing contextual cultural knowledge \cite{Gemeinhardt2024}. Although these systems achieved reliable responses, balancing creativity with scientific accuracy remains a key challenge, particularly in engaging young audiences with dynamic and accurate interactions.

The main goal of this research is to identify whether recent developments in frontier LLMs have allowed them to become suitable for Q\&A interactions with visitors in science centres. In order to do it, we collected a set of questions from visitors from a science centre, the National Space Centre in Leicester (UK), and prompted LLM models to provide responses. These responses were then assessed by science experts according to a set of criteria of scientific rigor and creativity. The main contributions of this paper are:
\begin{itemize}
    \item A method to analyse the trade-off between scientific rigor and creativity of frontier LLMs responses for Q\&A posed by visitors.
    \item A dataset of authentic questions frequently asked by visitors in a science centre, encompassing a broad range of inquiry types (closed, open, divergent, wildcard).
\end{itemize}

The remainder of this paper is structured as follows: Section 2 reviews related work on balancing creativity and factual accuracy in LLMs within educational contexts. Section 3 outlines the experimental setup, including dataset creation and evaluation methods. Section 4 presents the results, comparing model performance across key metrics. Finally, Section 5 discusses the findings and their implications for using LLMs in science centres, concluding with recommendations for future research.

\section{Related Work}

The rapid advancements in LLMs have opened new avenues for educational and cultural applications, particularly in science centres. These environments demand a careful balance between creativity, factual accuracy, and user engagement. This section reviews existing literature, moving from foundational research on balancing creativity and factuality in LLMs to applications in specialized contexts such as science tutoring and visitor attraction experiences.

\subsection{Balancing Creativity and Factual Accuracy in LLMs}

A critical aspect of applying LLMs is the trade-off between creative generation and factual accuracy, especially in settings that prioritize reliable information alongside engaging content. The approach in \cite{sinha2023mathematicalabstractionbalancingtradeoff} addressed this balance through a mathematical framework using a combined loss function, enabling fine-tuning that adjusts responses based on specific application needs. This adaptable approach provides a theoretical foundation for managing creativity and factuality in LLM outputs, setting a basis for applications in contexts such as interactive educational platforms and digital learning environments.

Further investigation into the effects of model alignment on creativity was conducted by \cite{mohammadi2024creativityleftchatprice}, who examined how Reinforcement Learning from Human Feedback (RLHF) impacts LLMs. The study revealed that RLHF reduces biases and improves response safety but can restrict creativity by forming “attractor states”, where responses become uniform. This phenomenon has implications for models used in engaging applications like Q\&A in science centres, where originality and accuracy must coexist to effectively captivate and inform visitors.

An additional approach to managing creativity within factual constraints was explored by \cite{jiang2024surveylargelanguagemodel}, who analyzed hallucinations in LLMs from a creative perspective. The authors suggested that, when controlled, hallucinations can enhance creativity by introducing innovative responses in scenarios that do not require strict factual adherence. This approach is relevant to settings such as science centres, where creative responses can foster visitor interest, provided that content remains broadly grounded in reality.

\subsection{Educational Applications of LLMs}

LLMs have demonstrated significant potential as educational tools, particularly in science and technology fields where accuracy and interactive learning are crucial. In \cite{chevalier2024languagemodelssciencetutors} the authors developed TUTOREVAL and TUTORCHAT, two benchmarks designed to evaluate and enhance the ability of LLMs to engage in structured science tutoring. These benchmarks show how LLMs can support accurate, interactive tutoring in STEM disciplines, establishing a framework for deploying LLMs in science centres, where educational rigor and engagement are paramount.

Beyond tutoring, recent research has employed LLMs to improve educational experiences through Question Answering (Q\&A) systems, particularly in science centres and cultural institutions. Notably, in \cite{bongini2023gpt3needvisualquestion}, the authors examined GPT-3’s performance in Visual Question Answering (VQA) for cultural heritage applications, where the model generated contextual descriptions of artworks. However, it struggled with detailed visual interpretations, indicating that while LLMs excel in contextual understanding, supplementary techniques may be necessary for fine-grained visual analysis in science centre settings.

\subsection{LLMs for Visitor Engagement and Personalization in Museums}

Recent advancements by \cite{electronics12183829} have enabled LLMs to create more engaging, personalized experiences for museum visitors. The authors developed a LLM-based recommendation system tailored to individual visitor interests, providing dynamic and immersive museum tours. This approach aligns closely with the needs of science centres, where unique, customized interactions enhance visitor experiences by responding to specific user interests and inquiries.

According to \cite{bdcc7030148}, the exploration of the use of ChatGPT as a museum guide emphasized the importance of human oversight to ensure factual accuracy. This study demonstrated the model's ability to adapt tone and content based on visitor demographics, a concept resonating with the goal of creating adaptable educational experiences in science centres.

Further developments in personalization by \cite{christakopoulou2023largelanguagemodelsuser} introduced a framework for tracking ''user interest journeys'' through LLMs. Utilizing clustering and fine-tuning, their model aligns recommendations with visitors' long-term goals. This level of personalization supports continuous engagement, proving highly applicable in science centres, where repeated interactions and sustained engagement are desirable.

\subsection{Techniques to Improve Factual Consistency and Creativity}

To improve factual accuracy in Q\&A applications, as described in \cite{allemang2024increasingllmaccuracyquestion}, researchers proposed a method that integrates ontologies for real-time verification of SPARQL queries in knowledge graph systems. This approach dynamically detects and corrects factual inaccuracies, emphasizing the role of semantic verification in contexts where accuracy is critical, such as in enterprise Q\&A and science centres.

For creativity enhancement, based on findings of \cite{lu2024llmdiscussionenhancingcreativity}, a role-playing framework was introduced, employing LLM-agent dialogues with diverse perspectives to promote original thinking. This collaborative approach fosters creativity in model responses and is relevant to science centres, where engaging visitors with novel perspectives is valuable. In \cite{zhao2024assessingunderstandingcreativitylarge}, LLM creativity was further assessed through adapted Torrance Tests, revealing that while LLMs display fluency and flexibility, they often lack originality. These insights are significant for science centres aiming to balance factuality with engaging responses that capture visitor interest.

According to \cite{bdcc7030148}, the exploration of the use of ChatGPT as a museum guide emphasized the importance of human oversight to ensure factual accuracy. This study demonstrated the model's ability to adapt tone and content based on visitor demographics, a concept resonating with the goal of creating adaptable educational experiences in science centres.

\subsection{Contributions and Positioning of Current Research}

Collectively, these studies illustrate the progress in balancing creativity, factual accuracy, and personalized engagement in LLMs, spanning applications from science education to cultural heritage. Building on this foundation, our study evaluates frontier models like GPT, Gemini, and Claude in the Q\&A context of science centres. We focus on determining their suitability for engaging young audiences while maintaining high standards of factual accuracy, specifically addressing the challenge of balancing educational rigor with engaging and creative responses tailored for visitor interactions.

\section{Experimental Setup}\label{setup}

In this section, we provide a detailed account of how the experiments were set up to assess the suitability of frontier LLMs to answer visitors' questions at science centres.  The National Space Centre (NSC) served as a case study, where we collected commonly asked questions from visitors. Responses were then generated by the chosen LLMs and assessed by science experts from the NSC.

\subsection{The National Space Centre}

The National Space Centre is the UK's largest space-themed visitor attraction, educating visitors in the excess of 300,000 per year. It contains a range of exhibits informing visitors on space science topics including the history of human space exploration, the science of celestial bodies and potential future developments in space travel. This information is currently conveyed through a wide range of interactive activities, displays and planetarium shows in the UK's largest planetarium. 

\subsection{Questions Dataset}

Initially, an internal review at the National Space Centre aimed to identify frequently asked questions, and three potential sources were recognized: i) questions from visitor polling, ii) recorded frequently asked questions that the Space Centre staff were trained to answer, and iii) questions submitted during 'Ask-the-Experts' events geared towards adult space enthusiasts. This approach ensured diversity in the dataset, catering to various visitor groups with different levels of specificity and prior knowledge. From the initial dataset of 141 questions (see Table \ref{tab:visitors_questions} in the Appendix for all questions), a subset of 12 was selected after eliminating duplicates and unsuitable entries. These questions were chosen to examine different response styles and aspects and were categorized into four groups with three questions each: closed, open, divergent, and wildcard questions. Closed questions require accurate responses whilst open questions allowed for an expert reviewer to interpret an answer with multiple facets, facts and statements. Divergent questions are related to predictions and can be evaluated by an expert. Finally, wildcard questions, touching on philosophical aspects, lack scientifically valid answers. In Table \ref{tab:questions}, we present all the questions used in this research.

\begin{table}[t]
    \centering
    \begin{tabular}{|m{4cm}|m{8cm}|}
        \hline
        \textbf{Category} & \textbf{Questions} \\
        \hline
        Closed Questions & How big is the moon? \\ 
        & How long does it take to get to space? \\ 
        & Is the Earth flat? \\
        \hline
        Open Questions & Why does NASA study the ocean? \\ 
        & What is the purpose of the golden disc on Voyager 1? \\ 
        & Why have we not gone back to the moon? \\
        \hline
        Divergent Questions & Do you think we will ever live on Mars? \\ 
        & Is food good in space?  \\ 
        & What is your favorite thing about space? \\
        \hline
        Wildcard & What is Space? Where does Space Start and End? \\ 
        & What is life like in space? \\ 
        & What is a white hole and how is it made? \\
        \hline
    \end{tabular}
    \caption{Questions per category. }
    \label{tab:questions}
\end{table}

\subsection{LLMs Responses Generation}

We selected the three most advanced frontier models to generate responses for the visitors' questions: OpenAI's GPT-4, Claude 3.5 Sonnet and Google Gemini 1.5. Each LLM was prompted to generate two responses for each question. In Table \ref{tab:promtps}, the two types of prompts are presented: \textbf{standard} and \textbf{creative}. The standard prompt demands short and concise answers suitable for an eight-year-old audience, a common age group among National Space Centre visitors due to the inclusion of astronomic topics in the British science syllabus for this age. Although the questions were posed from a diverse range of ages and sources, the science centre would require that a model be capable of fielding questions from adults and answering a manner that includes younger visitors therefore analysing against a target age for the models response allows for a review of the models suitability with the target market in mind. The creative prompt maintains this requirement but also encourages the model to produce responses that are both surprising and unconventional. This tests the LLM's ability to uphold high standards of accuracy and clarity while aligning with the science centre's secondary objective of engaging its audience.

\begin{table}[h!]
\centering

\begin{tabular}{|>{\raggedright\arraybackslash}p{3cm}|>{\raggedright\arraybackslash}p{8cm}|}
\hline
\textbf{Type} & \textbf{Prompt} \\ \hline
\textbf{standard} & Answer this question in 50 words or less. You are aiming to educate an 8-year-old. \\ \hline
\textbf{creative} & Answer this question in 50 words or less. You are aiming to educate an 8-year-old. Try to be creative and novel in your response. \\ \hline
\end{tabular}
\caption{Prompts used to generate \textbf{standard} and \textbf{creative} responses.}
\label{tab:promtps}
\end{table}

Models responded with significantly different answers when prompted with the creative prompt as opposed to the standard prompt with example including Claude's responses to the prompt 'How big is the moon?'. Claude responded with the statement 'The moon is very big, about the size of the United States. It looks small because it’s far away in space, around 238,000 miles from Earth. The moon is
the brightest object we see in the night sky' when asked the standard prompt and 'The moon is so big that if it were made of cheese, it could make enough pizza to feed everyone on Earth for over a million years! Its true size is about one-fourth the size of planet Earth.' when asked to provide a creative response. The standard prompt generated numerical measurements and the creative prompt generated a more abstract analogy. All the standard and creative responses generated by each model can be found in Tables \ref{tab:gpt_standard}, \ref{tab:gpt_creative}, \ref{tab:claude_standard}, \ref{tab:claude_creative}, \ref{tab:gemini_standard}, \ref{tab:gemini_creative} in the Appendix section.

\subsection{Science Experts Assessment}

The final stage of the experiments was to send the generated AI responses to the education teams at the National Space Centre, specifically targeting those identified as space experts. The review board of five experts had their status validated through a range of academic and professional qualifications including relevant PhDs and degrees in space science related fields, years of experience within the education sector and significant careers as communicators within the National Space Centre and at educational events across the world. Given their roles at the NSC, these experts are specifically focused on a high level of accuracy and clarity when educating visitors on interesting and engaging information about space and space science. A subset of responses, chosen for their varied outputs across models, was randomized and presented to the experts for evaluation using a five-point Likert scale. Responses were assessed in a blind testing environment, with original prompts included but without indicating which LLM generated each response, so to ensure unbiased assessment. Evaluation criteria included accuracy, clarity, engagement, deviation from typical expert answers, and level of surprise. Namely, for each response, each science expert was required to answer the 5 questions listed in Table \ref{tab:survey}.

\begin{table}[t]
\centering
\renewcommand{\arraystretch}{1.5} 
\begin{tabular}{|>{\raggedright\arraybackslash}p{5cm}|}
\hline
\textbf{Questions to Experts} \\ \hline
(a) Is this response accurate? \\ \hline
(b) Would this response clearly communicate to an 8-year-old child? \\ \hline
(c) Would this response engage an 8-year-old child? \\ \hline
(d) Is this response different from what I would normally say? \\ \hline
(e) Is this response surprising? \\ \hline
\end{tabular}
\caption{Survey questions to science centre experts to assess the LLMs' responses.}
\label{tab:survey}
\end{table}

\section{Results}\label{results}

\begin{figure}[t] 
    \centering
    \includegraphics[width=0.9\textwidth]{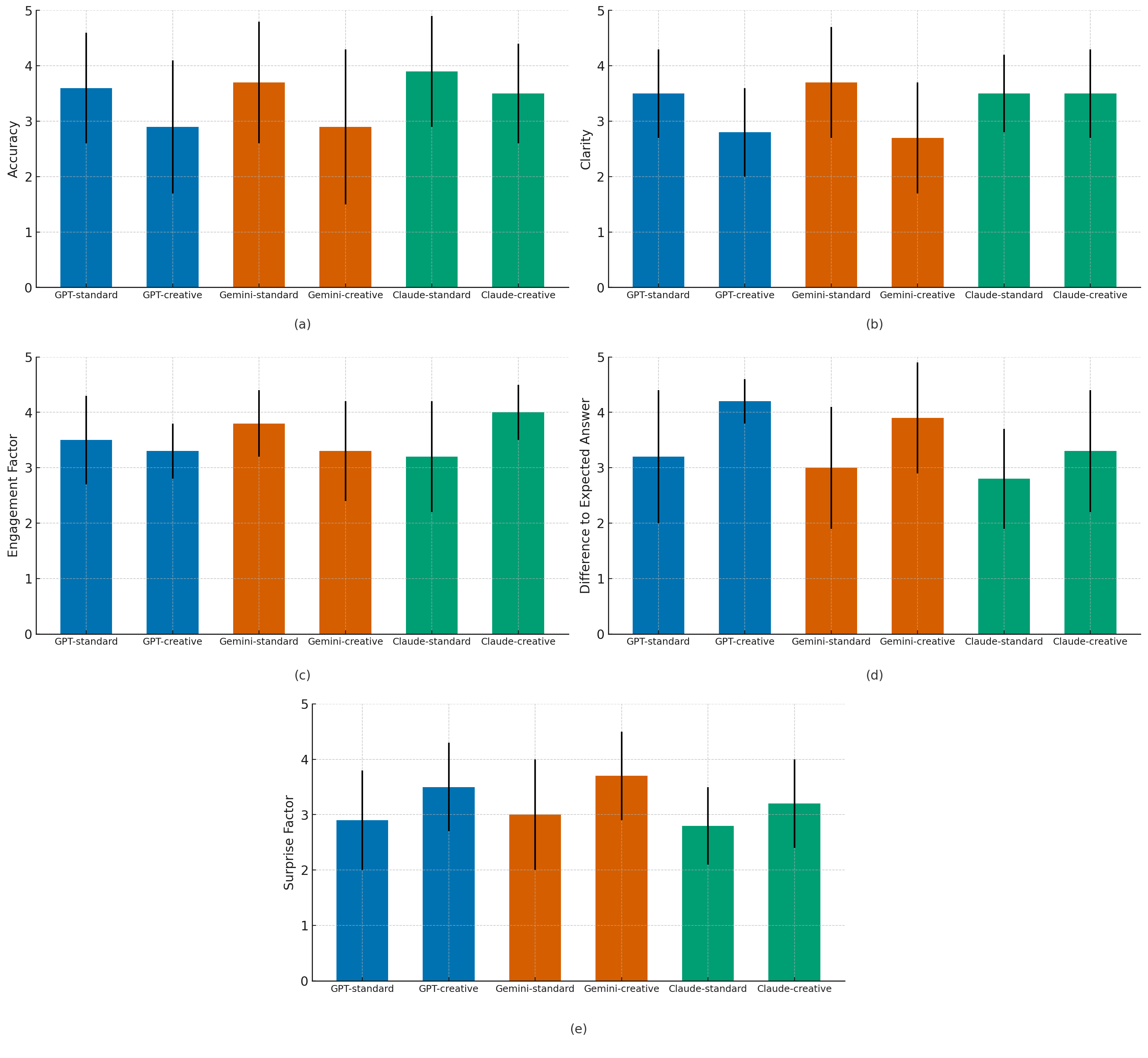} 
    \caption{Average scores assigned by science experts across five evaluation metrics: (a) accuracy, (b) clarity, (c) engagement, (d) deviation from expected answers, and (e) surprise. They compare the responses of GPT-4, Claude 3.5, and Gemini 1.5 under standard and creative prompts.}
    \label{fig:scores} 
\end{figure}

\begin{figure}[t] 
    \centering
    \includegraphics[width=0.9\textwidth]{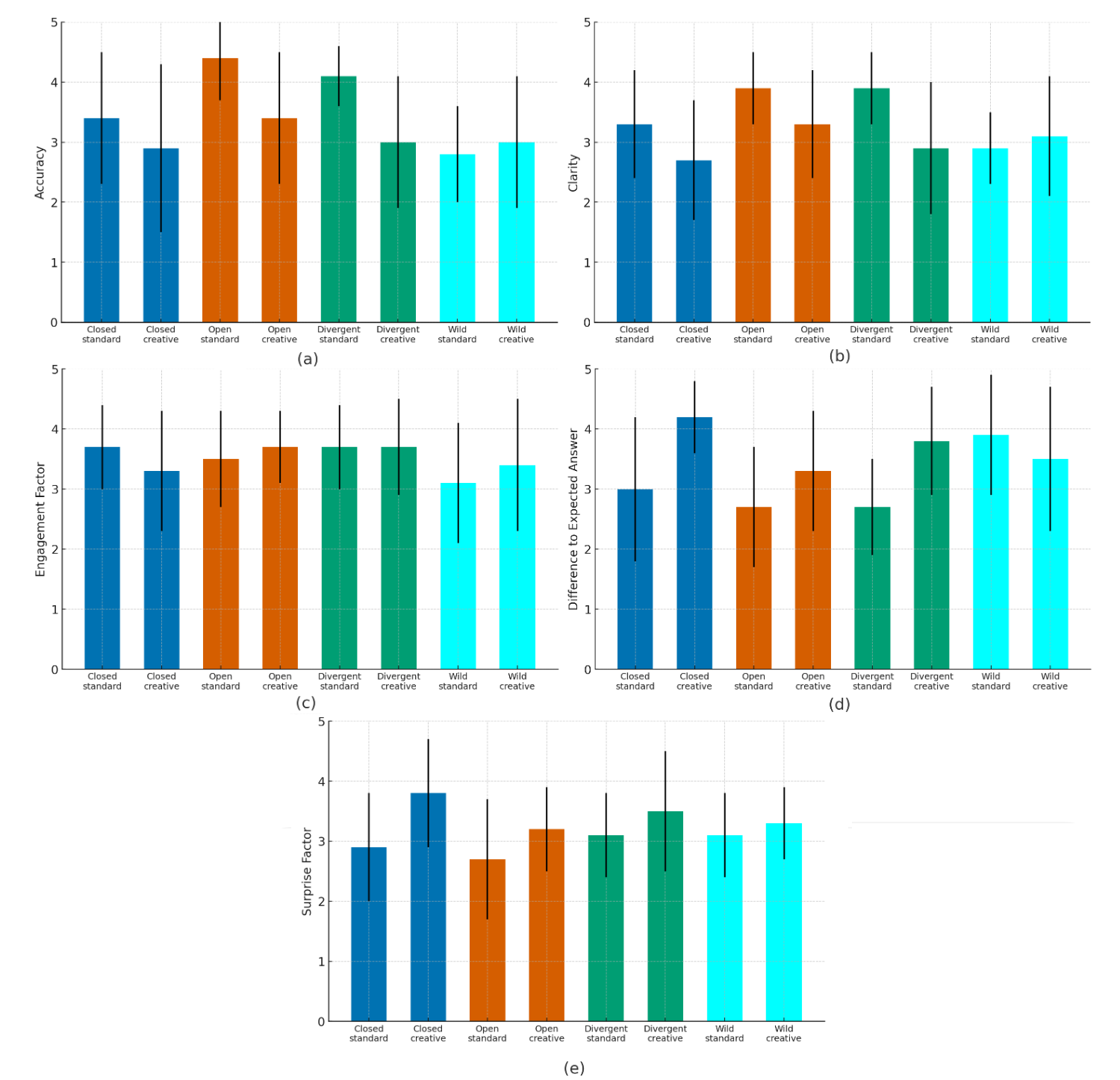} 
    \caption{Average scores for responses to closed, open, divergent, and wildcard questions, aggregated for all LLMs and categorized by standard and creative prompts. Scores highlight differences in  performance for the following metrics: (a) accuracy, (b) clarity, (c) engagement, (d) deviation from expected answers, and (e) surprise.}
    \label{fig:question scores} 
\end{figure}

In this section, we present the main results from our experiments. Figure \ref{fig:scores} presents a comparison of the average scores for the three selected LLMs (OpenAI’s \textbf{GPT}-4, \textbf{Claude} 3.5
Sonnet, and \textbf{Gemini} 1.5), while Figure \ref{fig:question scores} shows the average scores for each question type from combined LLM responses.

\subsection{Metrics Used}

Experts from the science centre rated the models based on the evaluation questions listed in Table \ref{tab:survey}, with scores ranging from 1 (lowest) to 5 (highest) for each LLM response. Each metric corresponds to a specific evaluation question. Error bars indicate the standard deviation of scores, illustrating the variability in each model’s performance. The metrics can be described as follows:

\begin{itemize}
    \item \textbf{(a) Accuracy}: This measures how well the models’ responses align with correct answers. Higher scores indicate greater accuracy.
    \item \textbf{(b) Clarity}: This evaluates the clarity of communication, assessing how understandable the response is for a young child. Higher scores indicate simpler, more accessible language.
    \item \textbf{(c) Engagement Factor}: This reflects how engaging each model’s response is for a young audience. Higher scores suggest that the response is likely to capture the interest of an 8-year-old child. 
    \item \textbf{(d) Difference to Expected Answer}: This measures the deviation from what an expert might typically explain. Lower scores indicate responses that are closer to an expert’s expected answer, while higher scores reflect greater deviation.
    \item \textbf{(e) Surprise Factor}: This represents the level of novelty or unexpectedness in the responses, indicating creativity or novelty. Higher scores suggest a more surprising answer.
    \end{itemize}

\subsection{Analysis per LLM}

Figure \ref{fig:scores}(a) shows that Claude consistently presented higher accuracy than the other models for both \textbf{standard} and \textbf{creative} responses. Claude also presented higher scores for clarity and engagement (Figures \ref{fig:scores}(b) and \ref{fig:scores}(c)) when prompted for creativity, with answer including 'NASA studies the ocean because Earth is a water world; oceans cover 71\% of our planet's surface! By studying the vast oceans from space, NASA can uncover secrets about Earth's climate, weather, and even the origins of life itself. Exploring the deep blue is key to understanding our amazing home planet.' in response to the prompt 'Answer this question in 50 words or less. You are aiming to educate an 8 year old. Try to be surprising and unexpected. Why does NASA study the ocean?'. Answers like this outperformed GPT and Gemini, which both exhibited declines in both accuracy and engagement with the creative prompt. Claude's average accuracy decreased by only 0.4, whereas both GPT and Gemini experienced significant reductions in their reviewed scores.

A similar trend is observed in the expert ratings for clarity. Claude's average clarity score remained unchanged when the creative prompt was used, while the scores for Gemini and GPT both decreased. Gemini providing responses including 'Astronauts might live on Mars someday, like explorers in a giant bug suit! But first, we gotta build a pizza place - gotta have fuel for all that exploring, right?' to the prompt 'Answer this question in 50 words or less. You are aiming to educate an 8 year old. Try to be surprising and unexpected. Do you think we will ever live on Mars?' which confused the experts and was reviewed with particularly low clarity. This suggests that Claude's LLM is more resilient to quality loss when attempting to be both educationally effective and creative in a science centre setting.
It is also interesting to note that the use of a prompt aiming at creativity did not actually increase the engagement level of GPT's and Claude's answers.

Figures \ref{fig:scores}(d) and \ref{fig:scores}(e) reveal that experts rate surprising responses, which deviate from their usual answers, as less accurate and clear. This reinforces the fact that, for science centre experts, accuracy cannot be compromised for engagement or creativity. It also reveals that current frontier models require further enhancements to achieve the high standards of science centres before being fully adopted for Q\&A purposes.

\subsection{Analysis per Question Type}

Figure \ref{fig:question scores} presents the results based on the same metrics as above, but organized by question types (closed, open, divergent, wildcard). This graph offers additional insights into the trade-offs between factual alignment and engagement with respect to the diverse types of queries visitors might pose and to how LLMs are prompted to generate responses. Figures \ref{fig:question scores}(a) and \ref{fig:question scores}(b) demonstrate that responses generated using standard prompts consistently achieve higher scores in accuracy and clarity, particularly for the open-question and divergent-question categories. This confirms, as noted in the previous section, that concise and straightforward prompts result in precise and easily comprehensible answers suitable for an 8-year-old audience. As expected, prompts aimed at fostering creativity exhibit a certain decline in accuracy and clarity, indicating that an emphasis on creativity may inadvertently introduce factual inaccuracies or complicate language. Conversely, we notice increased scores in Figures \ref{fig:question scores}(d) and \ref{fig:question scores}(e) when creative prompts are used; this might suggest that a high level of unexpectedness and surprise can sometimes simply arise from a factually incorrect answer.
A notable exception is observed in the wildcard category, where shifting from standard to creative prompts leads to a slight increase in accuracy and clarity, accompanied by a similar rise in the surprise factor. This can be explained by the inherent nature of wildcard questions, which lack an objectively valid answer.

Overall, our experimental findings suggest that standard prompts are more effective in achieving the primary goals of accuracy and accessibility in educational contexts. Conversely, creative prompts excel in metrics such as deviation from expected answers and the surprise factor. This emphasizes the importance of tailoring prompts to the specific educational or communicative goals in science outreach, and possibly to the type of questions being answered. For instance, while standard prompts seem to perform better for closed questions, where accuracy is crucial, our study suggests that prompts designed to induce creative answers can produce more appropriate responses for questions that encourage imaginative thinking, such as those in our wildcard category.

\section{Conclusion}

This study investigated the potential of frontier Large Language Models (LLMs) to generate responses that engage and educate young audiences in science centres, focusing specifically on 8-year-old children. Our findings indicate that Claude consistently outperformed GPT-4 and Gemini in maintaining educational quality, even when prompted to produce creative responses. This superior performance demonstrates that LLMs like Claude have significant potential to enhance educational experiences by effectively balancing engagement with factual accuracy. Our study highlights the necessity for careful prompt engineering and potential fine-tuning to mitigate losses in accuracy when aiming for increased creativity. The varying effectiveness of different prompt types across question categories underscores the importance of adaptive strategies in LLM deployment.

A limitation of this research is the reliance on expert assessments without direct feedback from the target young audience. Addressing this limitation, future work should involve refining prompt structures to better balance creativity and clarity across different models and expanding the assessment framework to include feedback from actual young audiences. Additionally, experimenting with adaptive prompt strategies based on the nature of the question, such as distinguishing between factual and conceptual queries, could further enhance the educational effectiveness of LLMs in science centres.

In conclusion, while LLMs hold great promise for educational applications, particularly in engaging young audiences, achieving the optimal balance between creativity and accuracy remains a challenge. By addressing this balance through targeted prompt engineering and model refinement, we can successfully integrate LLMs into educational settings like science centres, ultimately enriching the learning experience for young visitors and fostering a lifelong interest in science.

\section{Ethics} 
This research has been approved by the University of Leicester Ethics Committee under the title Creativity Evaluation Of Generative AI In Museum Exhibitions and Project ID 0696 and authorized by the National Space Centre.

\section{Acknowledgments} 
JW, FG and MV are supported by the University of Leicester (UK), in particular the Computing and Mathematical Sciences Department. TM is supported by UFOP (Federal University of Ouro Preto, Brazil). JW is also supported by the National Space Centre (NSC) Creative at Leicester (UK). As a charitable organization, the National Space Centre maintains a neutral stance and does not endorse any specific model over others. 

\section{Author contributions}
Experimental design: JW, FG, MV, TM; Implementation: JW; Writing and Editing: JW, FG, MV, TM.

\bibliographystyle{IEEEtran}
\bibliography{main}
\newpage
\section{Appendix}

\renewcommand{\arraystretch}{1.5} 

\begin{longtable}{|>{\raggedright\arraybackslash}p{4cm}|>{\raggedright\arraybackslash}p{11cm}|}

\hline
\multicolumn{2}{|c|}{\textbf{GPT}} \\ \hline
\hline
\multicolumn{2}{|c|}{\textbf{Answer this question in 50 words or less. You are aiming to educate an 8 year old.}} \\ \hline
\hline
\textbf{Prompt} & \textbf{Response} \\ \hline
\endfirsthead
\hline
\textbf{Prompt} & \textbf{Response} \\ \hline
\endhead
\hline
\endfoot

\multicolumn{2}{|c|}{\textbf{Closed Questions}} \\ \hline

How big is the moon? & The moon is pretty big, but not as big as Earth. It's about as wide as the United States from coast to coast. That's around 2,159 miles across. Imagine if you could drive a car all the way across the moon! \\ \hline

How long does it take to get to space? & To get to space, it takes only about 10 minutes by rocket. It's like a super-fast car going straight up into the sky until it's way above the clouds and Earth, entering what we call space. \\ \hline

Is the Earth Flat? & No, the Earth is not flat. It is shaped like a sphere, much like a round ball. So even if you walked and walked in a straight line, you'd eventually end up back where you started, because it's round, not flat like a pancake. \\ \hline

\multicolumn{2}{|c|}{\textbf{Open Questions}} \\ \hline

Why does NASA study the ocean? & NASA studies the ocean to understand how it affects our planet’s weather and climate. They use satellites to watch things like ocean temperature and sea levels. This helps scientists learn how to protect animals and people from environmental changes. \\ \hline

What is the purpose of the golden disc on Voyager 1? & The golden disc on Voyager 1 is like a space message in a bottle. It has pictures, sounds, and music from Earth. It's there to show aliens what life is like here, in case they find it floating in space far away. \\ \hline

Why have we not gone back to the moon? & We haven't gone back to the moon recently because it's very expensive and we learned a lot the first times we went. Now, scientists are studying other things in space, like Mars, and using robots and satellites to learn more about the moon and other planets from a distance. \\ \hline

\multicolumn{2}{|c|}{\textbf{Divergent Questions}} \\ \hline

Do you think we will ever live on Mars? & Yes, scientists are working hard to make it possible for people to live on Mars one day! They're studying Mars and creating new technologies to help humans live there safely. It might happen when you're much older, but it’s a big, exciting possibility for the future! \\ \hline

Is food good in space? & In space, astronauts eat different kinds of food that are packed specially to stay good. They might not be as tasty as food on Earth because there’s no fridge or oven, but they still have all the nutrients to keep astronauts healthy and strong while they’re away from home. \\ \hline

What is your favorite thing about space? & My favorite thing about space is that it's full of amazing mysteries! There are stars, planets, and even black holes, which are like cosmic vacuum cleaners. Also, space is so big that there are secrets we haven't even discovered yet! Imagine, there might be other planets out there with alien life. \\ \hline

\multicolumn{2}{|c|}{\textbf{Wildcard Questions}} \\ \hline

What is Space? Where does Space Start and End? & Space is like a huge, endless playground where planets, stars, and galaxies float. It starts where Earth's air gets too thin to breathe, far above the clouds. Nobody knows where space ends, it keeps going on and on! \\ \hline

What is life like in space? & Life in space is like camping in a very special tent. Astronauts float around because there's no gravity, eat special packaged food, and sleep in bags hooked to the wall. They also do a lot of science experiments to learn how things work without gravity! \\ \hline

What is a white hole and how is it made? & A white hole is like the opposite of a black hole. Instead of sucking things in, it pushes things out. Scientists think it could be made when a black hole stops pulling and starts pushing, but no one has seen a white hole yet. It's still a mystery! \\ \hline
\caption{GPT standard responses.}
\label{tab:gpt_standard}
\end{longtable}

\begin{longtable}{|>{\raggedright\arraybackslash}p{4cm}|>{\raggedright\arraybackslash}p{11cm}|}
\hline
\multicolumn{2}{|c|}{\textbf{GPT-Creative}} \\ \hline
\hline
\multicolumn{2}{|c|}{\textbf{Answer this question in 50 words or less. You are aiming to educate an 8 year old.}} \\ 
\multicolumn{2}{|c|}{\textbf{ Try to be surprising and unexpected.}} \\
\hline
\hline
\textbf{Prompt} & \textbf{Response} \\ \hline
\endfirsthead
\hline
\textbf{Prompt} & \textbf{Response} \\ \hline
\endhead
\hline
\endfoot

\multicolumn{2}{|c|}{\textbf{Closed Questions}} \\ \hline

How big is the moon? & Imagine 50 moons lined up next to each other — that’s how long it would take to stretch across the United States from coast to coast! The moon is pretty big, almost like a quarter the size of Earth. Cool, right? \\ \hline

How long does it take to get to space? & Did you know it takes less time to get to space than to watch a movie? A rocket can blast you into space in just about 10 minutes! That's quicker than driving to your friend's house sometimes! Isn't that super fast? \\ \hline

Is the Earth Flat? & No, the Earth isn't flat; it's shaped like a giant sphere, like a basketball. Long ago, people even sailed ships all the way around it to prove this! If Earth were flat, we’d find the edge, but no one ever has because it’s a big, round ball spinning in space! \\ \hline

\multicolumn{2}{|c|}{\textbf{Open Questions}} \\ \hline

Why does NASA study the ocean? & NASA studies the ocean from space to understand how Earth's water and air interact, which affects our weather and climate. By looking at the ocean from above, they can see things we can't, like tiny changes that tell us important secrets about our planet’s health! Cool, right? \\ \hline

What is the purpose of the golden disc on Voyager 1? & The golden disc on Voyager 1 is like a space treasure map and hello note combined! It's made so that if aliens find it, they can learn about us with sounds and pictures from Earth. Imagine aliens discovering our space message and learning about our world! \\ \hline

Why have we not gone back to the moon? & We haven't gone back to the moon recently because we've been busy studying other exciting space things like Mars and distant galaxies! Now, scientists are planning new adventures to the moon to build bases and learn even more. It’s like planning a very cool space camp on the moon! \\ \hline

\multicolumn{2}{|c|}{\textbf{Divergent Questions}} \\ \hline

Do you think we will ever live on Mars? & Yes, we might live on Mars one day! Scientists and astronauts are working hard to figure out how to make Mars safe and cozy for people. Imagine playing space hopscotch or celebrating your birthday on a whole different planet! Isn’t that exciting and a bit wild? \\ \hline

Is food good in space? & Yes, food in space can be yummy, but it's different! Imagine squeezing your dinner from a tube like toothpaste or eating snacks that float like magic bubbles! Astronauts sometimes even get special treats like ice cream that's freeze-dried, so it's crunchy instead of cold and creamy! \\ \hline

What is your favorite thing about space? & My favorite thing about space is that it's full of astonishing mysteries, like black holes, which are so strong they can even suck up light! Imagine a cosmic vacuum cleaner so powerful, not even the fastest thing in the universe can escape it! Cool, right? \\ \hline

\multicolumn{2}{|c|}{\textbf{Wildcard Questions}} \\ \hline

What is Space? Where does Space Start and End? & Space is like an enormous, never-ending playground that starts where Earth's sky finishes. It has planets, stars, and even galaxies! We don't really know where it ends—it's like trying to find the edge of a coloring book that keeps adding pages! Isn't that cool? \\ \hline

What is life like in space? & Imagine floating around like a superhero every day because there’s no gravity in space! You get to sleep stuck to the wall and use special toothpaste so it doesn’t float away. Space food comes in cool packets, and you have to exercise lots so your muscles stay strong! \\ \hline

What is a white hole and how is it made? & Imagine if a black hole was a vacuum cleaner sucking everything in, a white hole is like a magic bubble blower that spits everything out! We're not sure if they exist, but scientists think they might be the opposite of black holes, popping out stuff instead of gobbling it up! \\ \hline
\caption{GPT creative responses.}
\label{tab:gpt_creative}
\end{longtable}

\begin{longtable}{|>{\raggedright\arraybackslash}p{4cm}|>{\raggedright\arraybackslash}p{11cm}|}
\hline
\multicolumn{2}{|c|}{\textbf{Claude}} \\ \hline
\hline
\multicolumn{2}{|c|}{\textbf{Answer this question in 50 words or less. You are aiming to educate an 8 year old.}} \\ \hline
\hline
\textbf{Prompt} & \textbf{Response} \\ \hline
\endfirsthead
\hline
\textbf{Prompt} & \textbf{Response} \\ \hline
\endhead
\hline
\endfoot

\multicolumn{2}{|c|}{\textbf{Closed Questions}} \\ \hline

How big is the moon? & The moon is very big, about the size of the United States. It looks small because it's far away in space, around 238,000 miles from Earth. The moon is the brightest object we see in the night sky. \\ \hline

How long does it take to get to space? & It takes about 8 minutes and 42 seconds for a powerful rocket to launch a spacecraft into space from Earth. The rocket burns fuel to lift the spacecraft out of Earth's atmosphere, escaping Earth's gravitational pull. \\ \hline

Is the Earth Flat? & No, the Earth is not flat. It is spherical, like a big ball. When you look at pictures taken from space, you can clearly see that the Earth is round and not flat. Scientists have proven this through many observations and experiments over centuries. \\ \hline

\multicolumn{2}{|c|}{\textbf{Open Questions}} \\ \hline

Why does NASA study the ocean? & NASA studies the oceans to understand Earth's climate system and how ocean conditions impact weather patterns, sea levels, and marine life. By monitoring oceans from space, NASA can track changes over time and help us better care for our planet's precious waters. \\ \hline

What is the purpose of the golden disc on Voyager 1? & The golden disc on Voyager 1 is a message to any intelligent life that might find it. It contains images, sounds, and messages from Earth, showing what our planet and civilizations are like. \\ \hline

Why have we not gone back to the moon? & We have not gone back to the moon recently because space exploration is very expensive and difficult. However, many countries are now working on new missions to study the moon more closely and even establish bases there in the future. \\ \hline

\multicolumn{2}{|c|}{\textbf{Divergent Questions}} \\ \hline

Do you think we will ever live on Mars? & Yes, I think humans will eventually live on Mars. It will take many years of hard work and technological advances, but Mars could provide a new home for humanity in the future as resources on Earth become strained. Living on Mars would be very difficult, but also incredibly exciting and pioneering. \\ \hline

Is food good in space? & Yes, food in space is good! Astronauts eat specially prepared meals that are nutritious and flavorful. However, without gravity, liquids can spill and float away, so astronauts have to be careful when eating and drinking in space. \\ \hline

What is your favorite thing about space? & My favorite thing about space is the vast, unexplored mysteries it holds. From distant galaxies to undiscovered planets, space is an endless frontier waiting to be explored and understood. The wonders of the cosmos never cease to amaze and inspire curiosity. \\ \hline

\multicolumn{2}{|c|}{\textbf{Wildcard Questions}} \\ \hline

What is Space? Where does Space Start and End? & Space is the vast emptiness that exists beyond Earth's atmosphere. It starts a few hundred miles above Earth's surface and extends infinitely in all directions, containing planets, stars, galaxies, and everything else in the universe. \\ \hline

What is life like in space? & Life in space is very different from Earth. There is no air to breathe, so astronauts need special suits. Everything floats because there is little gravity. Eating, sleeping, and using the bathroom require special techniques. The view of Earth and stars is breathtaking from space. \\ \hline

What is a white hole and how is it made? & A white hole is the theoretical opposite of a black hole. It is hypothesized to be a region in spacetime that cannot be entered from the outside, but from which matter and light can escape. White holes are thought to be created during the Big Bang. \\ \hline
\caption{Claude standard responses.}
\label{tab:claude_standard}
\end{longtable}

\begin{longtable}{|>{\raggedright\arraybackslash}p{4cm}|>{\raggedright\arraybackslash}p{11cm}|}
\hline
\multicolumn{2}{|c|}{\textbf{Claude-Creative}} \\ \hline
\hline
\multicolumn{2}{|c|}{\textbf{Answer this question in 50 words or less. You are aiming to educate an 8 year old.}} \\ 
\multicolumn{2}{|c|}{\textbf{ Try to be surprising and unexpected.}} \\
\hline
\hline
\textbf{Prompt} & \textbf{Response} \\ \hline
\endfirsthead
\hline
\textbf{Prompt} & \textbf{Response} \\ \hline
\endhead
\hline
\endfoot

\multicolumn{2}{|c|}{\textbf{Closed Questions}} \\ \hline

How big is the moon? & The moon is so big that if it were made of cheese, it could make enough pizza to feed everyone on Earth for over a million years! Its true size is about one-fourth the size of planet Earth. \\ \hline

How long does it take to get to space? & It takes about 9
minutes to reach space from Earth! Imagine being in a rocket so powerful, it can zip you through the sky faster than a speeding race car, leaving our planet behind. Space is an incredible journey waiting to be explored! \\ \hline

Is the Earth Flat? & No, the Earth is not flat! It's actually a huge spherical ball that spins and orbits the Sun. If you look at pictures taken from space, you can clearly see the Earth's curved shape. Isn't that amazing that we live on a gigantic, rotating sphere floating through the cosmos? \\ \hline

\multicolumn{2}{|c|}{\textbf{Open Questions}} \\ \hline

Why does NASA study the ocean? & NASA studies the ocean because Earth is a water world; oceans cover 71\% of our planet's surface! By studying the vast oceans from space, NASA can uncover secrets about Earth's climate, weather, and even the origins of life itself. Exploring the deep blue is key to understanding our amazing home planet. \\ \hline

What is the purpose of the golden disc on Voyager 1? & The golden disc on Voyager 1 is a time capsule with pictures, sounds, and messages from Earth for any intelligent aliens who might find it in the distant future. It's like a hello from humans to potential new friends among the stars! \\ \hline

Why have we not gone back to the moon? & It's puzzling, isn't it? We haven't gone back to the Moon because space exploration is really expensive and difficult. But maybe one day soon, we'll discover something amazing on the Moon that will make us want to visit again! Who knows what mysteries await us up there? \\ \hline

\multicolumn{2}{|c|}{\textbf{Divergent Questions}} \\ \hline

Do you think we will ever live on Mars? & Yes, we may live on Mars one day! Humans are working hard to explore and understand the Red Planet. With special suits, homes, and even farms, Mars could become a new home for adventurous space explorers. It would be an exciting frontier to discover! \\ \hline

Is food good in space? & Food in space is different from Earth! Astronauts eat special dehydrated foods that don't make crumbs. They also eat foods like tortillas that won't float away. Sometimes, they get fun treats like ice cream! Eating in space requires using straps to keep the food from floating off. \\ \hline

What is your favorite thing about space? & My favorite thing about space is the mystery of black holes - regions so dense that not even light can escape their gravity! I imagine black holes could be portals to other strange dimensions and worlds we can't yet comprehend. Space is full of wonders waiting to be discovered! \\ \hline

\multicolumn{2}{|c|}{\textbf{Wildcard Questions}} \\ \hline

What is Space? Where does Space Start and End? & Space is the vast, mysterious beyond that surrounds our planet Earth. It starts just a few miles above where planes fly, and extends outward forever without any limits or boundaries. Space holds countless wonders like twinkling stars, whirling galaxies, and maybe even alien worlds waiting to be explored! \\ \hline

What is life like in space? & Life in space is an incredible adventure! Imagine floating like a butterfly, eating food from a tube, and using a special toilet. You'd get to see our beautiful Earth from above and even go for spacewalks outside the spacecraft. Space is a magical place full of mysteries waiting to be explored! \\ \hline

What is a white hole and how is it made? & A white hole is the opposite of a black hole. Instead of sucking everything in, it spits out matter and energy! Scientists think white holes may have formed during the Big Bang that created our universe, spewing out all the galaxies, stars, and planets we see today. \\ \hline
\caption{Claude creative responses.}
\label{tab:claude_creative}
\end{longtable}

\begin{longtable}{|>{\raggedright\arraybackslash}p{4cm}|>{\raggedright\arraybackslash}p{11cm}|}
\hline
\multicolumn{2}{|c|}{\textbf{Gemini}} \\ \hline
\hline
\multicolumn{2}{|c|}{\textbf{Answer this question in 50 words or less. You are aiming to educate an 8 year old.}} \\ \hline
\hline
\textbf{Prompt} & \textbf{Response} \\ \hline
\endfirsthead
\hline
\textbf{Prompt} & \textbf{Response} \\ \hline
\endhead
\hline
\endfoot

\multicolumn{2}{|c|}{\textbf{Closed Questions}} \\ \hline

How big is the moon? & Imagine you could shrink Earth down to the size of a giant beach ball. The moon would be like a bouncy ball next to it! It's much smaller than Earth, but still really big, the fifth biggest moon in the whole solar system! \\ \hline

How long does it take to get to space? & It depends where in space you're going! To reach the International Space Station, astronauts zip up in just a few hours. But for faraway places like the Moon, it can take days! \\ \hline

Is the Earth Flat? & Nope! Earth is actually a giant ball, like a bouncy globe. That's why ships disappear hull first over the horizon, not shrinking smaller and smaller. \\ \hline

\multicolumn{2}{|c|}{\textbf{Open Questions}} \\ \hline

Why does NASA study the ocean? & NASA studies Earth's giant ocean to understand how it affects our weather, like giant waves in the sky! The ocean also keeps our planet healthy by soaking up extra heat. Cool, right? \\ \hline

What is the purpose of the golden disc on Voyager 1? & Imagine a message in a bottle, but for space! The golden disc on Voyager 1 is like a recording of Earth - sounds, pictures, and greetings - in case any friendly aliens ever find it, floating way out in space. It's a way to say hello from us! \\ \hline

Why have we not gone back to the moon? & We went to the moon a long time ago, kind of like going to a really cool park! It was super expensive, and after a while, everyone wanted to explore other things in space. But guess what? We're actually planning to go back soon, like visiting that cool park again with new things to discover! \\ \hline

\multicolumn{2}{|c|}{\textbf{Divergent Questions}} \\ \hline

Do you think we will ever live on Mars? & Living on Mars is tricky! It's cold with thin air, so we'd need special suits to go outside. But scientists are working on ways to grow food and make water there. Maybe one day, astronauts could visit for a long time, or even stay forever! \\ \hline

Is food good in space? & Space food isn't bad, but it's different! Astronauts eat healthy meals like fruits, veggies, and mac \& cheese, but they come in pouches or freeze-dried to last long journeys and float in zero gravity! It's like camping food for space adventures! \\ \hline

What is your favorite thing about space? & My favorite thing about space is all the surprises! There are swirling colorful clouds, giant balls of gas like Jupiter, and maybe even planets with friendly aliens! Space is a giant mystery waiting to be explored! \\ \hline

\multicolumn{2}{|c|}{\textbf{Wildcard Questions}} \\ \hline

What is Space? Where does Space Start and End? & Imagine Earth without air, that's space! It's a giant, dark vacuum filled with stars and planets. Space starts about 100km up, where the air gets thin, and goes all the way out, pretty much forever! \\ \hline

What is life like in space? & Imagine floating like a balloon! In space, there's no gravity so astronauts float and use straps to stay put. They eat space food in pouches and see amazing views of Earth! It's fun but also tricky, like brushing your teeth without gravity! \\ \hline

What is a white hole and how is it made? & A white hole is like a giant fountain in space, shooting out light and energy. Unlike black holes that gobble things up, scientists aren't sure if they even exist! They might be the other end of a black hole in a different universe, but that's just a wild guess! \\ \hline
\caption{Gemini standard responses.}
\label{tab:gemini_standard}
\end{longtable}

\begin{longtable}{|>{\raggedright\arraybackslash}p{4cm}|>{\raggedright\arraybackslash}p{11cm}|}
\hline
\multicolumn{2}{|c|}{\textbf{Gemini-Creative}} \\ \hline
\hline
\multicolumn{2}{|c|}{\textbf{Answer this question in 50 words or less. You are aiming to educate an 8 year old.}} \\ 
\multicolumn{2}{|c|}{\textbf{ Try to be surprising and unexpected.}} \\
\hline
\hline
\textbf{Prompt} & \textbf{Response} \\ \hline
\endfirsthead
\hline
\textbf{Prompt} & \textbf{Response} \\ \hline
\endhead
\hline
\endfoot

\multicolumn{2}{|c|}{\textbf{Closed Questions}} \\ \hline

How big is the moon? & Imagine fitting all of the United States across the moon! That's right, our giant moon is like a cosmic playground, almost as wide as all of the USA! \\ \hline

How long does it take to get to space? & It depends! To reach the super cool International Space Station, it's like a car ride - just 6 hours to 3 days! But to touch the moon's dusty surface, it's like a long bike trip - 3 whole days! \\ \hline

Is the Earth Flat? & Nope! Earth is shaped like a giant marble, but smoother! Ships disappear hull first over the curve, just like your toy car going over a basketball. Cool, right? \\ \hline

\multicolumn{2}{|c|}{\textbf{Open Questions}} \\ \hline

Why does NASA study the ocean? & NASA isn't just looking for aliens in spaceships! They also search for them in giant, cosmic oceans hiding beneath the ice on moons of Jupiter and Saturn! Pretty cool, right? \\ \hline

What is the purpose of the golden disc on Voyager 1? & Voyager 1 is like a tiny time capsule flying through space! The golden disc is a message for any aliens who find it, filled with pictures and sounds of Earth, like greetings in different languages and funny animal noises! It's like a postcard from our whole planet! \\ \hline

Why have we not gone back to the moon? & Astronauts are busy! After reaching the moon, they realized there's a whole galaxy to explore, like deep sea divers finding a whole new ocean under the waves! We're zooming around the solar system now, but the moon will always be there for a return visit! \\ \hline

\multicolumn{2}{|c|}{\textbf{Divergent Questions}} \\ \hline

Do you think we will ever live on Mars? & Astronauts might live on Mars someday, like explorers in a giant bug suit! But first, we gotta build a pizza place - gotta have fuel for all that exploring, right? \\ \hline

Is food good in space? & Space food can be surprising! It comes in pouches, cubes, and even looks like astronaut ice cream! It might not taste like pizza, but it fuels your space adventures! \\ \hline

What is your favorite thing about space? & My favorite thing about space is that there might be planets made of giant ice cream scoops! Imagine swimming in a chocolate river on a gummy worm planet! Pretty cool, right? \\ \hline

\multicolumn{2}{|c|}{\textbf{Wildcard Questions}} \\ \hline

What is Space? Where does Space Start and End? & Space is like a giant invisible ocean, filled with twinkling stars like glitter! It starts just above Earth's thin atmosphere, kind of like the edge of a swimming pool, and keeps going forever, as far as we can imagine! \\ \hline

What is life like in space? & In space, you float like a superhero! But brushing your teeth is tricky - bubbles fly everywhere! You see amazing views of Earth, like a giant blue marble, and watch incredible sunrises every 90 minutes! \\ \hline

What is a white hole and how is it made? & Imagine a giant space fountain spraying out fireworks! That's kind of like a white hole, a super mysterious place that shoots out light and energy instead of sucking it in. Scientists aren't sure they exist, but maybe they're made from leftover stardust after a super big bang! \\ \hline
\caption{Gemini creative responses.}
\label{tab:gemini_creative}
\end{longtable}

\newpage

\begin{longtable}{|p{0.3\textwidth}|p{0.6\textwidth}|}
\hline
\textbf{Category} & \textbf{Questions} \\ 
\hline
\endfirsthead
\hline
\textbf{Category} & \textbf{Questions} \\ 
\hline
\endhead
\hline
\endfoot

\hline
\textbf{Space Crew Feedback} & Are the lights on the ISS LEDS or are they traditional bulbs? \newline
What is the weight/mass of the piece of the moon? \newline
Was the moon rock analyzed on what type of rock it is, is it volcanic? \newline
Why is planet Earth where it is? \newline
Where was Black Arrow launched from? \newline
Why was the Space Centre built in Leicester? \newline
Why have we not gone back to the moon? \newline
Is there water on the moon and Mars, and why is it important? \newline
Are we going to build a replacement ISS to the current one? \newline
Is the city in the timelapse projected in ‘Home Planet’ based on a real city? \newline
Are the rocks on the ‘Rock Table’ naturally made or man-made? \newline
Why is Saturn in the bathtub in our Solar System Gallery? \newline
How far away is the moon from Earth? \newline
Does the Sun orbit the Earth and is it projected? \newline
How does the Sun not move and how does it keep us rotating? \newline
What is Space? Where does Space start and end? \newline
Are the two rockets in the rocket tower real? \newline
What role does the UK have in current Space missions \newline and in the Space industry? \newline
Is the Earth flat? \newline
Does the UK have its own fleet of rocket systems to send people into space? \newline
What are the latest instruments sent into space by the Uni of Leicester? \newline
How easy is it to become an astronaut? \newline
Can visitors go into space from the Space Centre? \newline
How big is the moon? \newline
What are the sparkly bits in the piece of the Moon? \newline
What is the size of the planets in relation to one another? \newline
Why are some planets big and gassy while others are small and rocky? \newline
Do other planets have a core? \newline
Can you surf on the rings of Neptune like in the movie *Ad Astra*? \\

\hline
\textbf{Space Staff Feedback} & What the National Space Centre is built from? \newline
What is this? Does it cost anything? \newline Will I be waiting 25 minutes? Is it like the planetarium show? \newline
Why haven't we been back to the Moon since Apollo? \newline
How the National Space Centre was built? \newline
Where/what is my show? (Planetarium show) \newline
What is a black hole and how is it made? Same for white holes. \newline
How they got the rockets inside the Space Centre? \newline
What here has actually been to space? \newline
Why are the constellations what they are? \newline
Some questions about the James Webb telescope. \\

\hline
\textbf{Ask the Expert Questions} & What makes the rockets go up into space? \newline
What does it feel like when you blast off? \newline
What does it feel like when you land? \newline
What training is needed to become an astronaut? \newline
Why do astronauts float? \newline
How small are the dwarf planets Eris and Pluto? \newline
What food do astronauts eat in space? How do astronauts eat in space? \newline What does an astronaut kitchen look like? \newline Why do they eat dried, hard strawberries and not fresh juicy ones? \newline
How do astronauts drink in space? \newline
How do astronauts have a bath? \newline
Why do they have a seatbelt on to sleep? \newline
How do you keep fit and healthy in space? \newline
Why do astronauts wear a helmet? How warm is it in space? \newline What does the wire do linked to an astronaut \newline when they are out of the space craft? \newline
What can you see out of the window in space? \newline
What is life like in space? \newline
How was space created? \newline
What is the lifespan of a star? \newline
What are nebulas and what are they made of? \newline
What is your favorite thing about space? \newline
How are black holes created? \newline
What happens if you enter a black hole? \newline
Are you aware of any life in space? \newline
Is there another planet out of the solar system that has a magnetic field? \newline
Do any planets have active volcanoes? \newline
What is the coldest planet? \newline
Do you think we will ever live on Mars? \newline
Is there a reason why people have gone to Mars but not Venus? \newline
What’s the pressure like in space? Is it similar to the ocean? \newline
Why does space not have gravity like Earth? \newline
Is there internet in space? Can you play video games? \newline
Do you know anyone who went to space? \newline
How do astronauts exercise in their ships? \newline
Is food good in space? What do astronauts eat? \newline
How do you poop in space and what do you do with the remains? \newline
Do you get younger in space? \newline
What made you interested in space? \newline
Do you get the internet in space? \newline
In space, how do you get medical attention? \newline
How does it feel in space and how does it feel when you return? \newline
Why does NASA study the ocean? \newline
What asteroids have hit Earth? \newline
What temperature is the sun and what is it made of? \newline
How many stars are in space? \newline
How does a black hole work and could you get close to one? \newline
Why was Pluto changed to a dwarf planet? \newline
How is the atmosphere formed? \newline
Are we able to live on another planet? \newline
When will we go to Mars and how long would it take to get there? \newline
What is the purpose of the golden disc on Voyager 1? \newline
What would happen if you found extra-terrestrial life? \newline
What is the most beautiful thing you can see in space? \newline
How long does it take to get to space? \newline
What does it feel like in the rocket? Is it comfortable? \newline
How many people can fit into the rocket? / On the ISS? \newline
How do you steer / navigate the rocket? \newline \\

\textbf{Ask the Expert Questions} & What kind of special training do you have to do? \newline
How long can you stay in space for at one time? \newline
Where and how do you sleep—what is it like? \newline
What do you eat and drink and how? \newline
How does it feel inside the space suit? \newline
How much oxygen do you have in your spacesuit? \newline
Does it feel cold up in space? \newline
What sort of experiments do you do? \newline
What is it like looking at outer space / in orbit—what can you see? \newline
Can you space walk on different planets? \newline
How do you keep clean in space? \newline
How do you go to the toilet in space? \newline
How do you stay in touch with people back on Earth? \newline
Is it a hard job to do? \newline
What is it like to be an astronaut? \newline
Why do you choose to go to space? \newline
How does a rocket work? \newline
How long does it take to get to space? \newline
What training do you need to do to be an astronaut? \newline
How do you become an astronaut? \newline
Why do you float in space? \newline
How long are you in space for? \newline
What do you wear in space? \newline
What do astronauts eat in space? \newline
What is a black hole? \newline
Which planets have you seen? \newline
What does the Earth look like from space? \newline
How do you have a shower in space? \newline
How do you go to the toilet in space? \newline
Where do you sleep in the rocket? \newline
Is it scary in space? \newline
Why does the Moon sometimes look bigger or smaller? \newline
Is there always someone in space? \newline
What powers the International Space Station? \newline
Is there lots of rubbish in space and is this a problem? \newline
What happens if you are in space without protective gear? \newline
Are there medical conditions associated with space travel? \newline
What makes space so cold? \newline
When is space considered space and not the sky? \newline
What is a black hole? \newline
How environmentally-friendly is space travel? \newline
How much does one rocket launch cost? \newline
How close are we to a flight to Mars? \newline
How do you go to the toilet and keep clean in space? \newline
What is the strangest thing about being in space? \newline
How do astronauts entertain themselves? \\
\hline

\caption{Visitors questions.}
\label{tab:visitors_questions}

\end{longtable}

\end{document}